\documentclass{article}
\pdfoutput=1
\usepackage{spconf,amsmath,graphicx,amssymb}
\usepackage{cite}
\usepackage{textcomp}
\usepackage{stfloats}
\usepackage{url}
\usepackage{booktabs}
\usepackage{color}
\usepackage[caption=false]{subfig}
\usepackage[colorlinks,linkcolor=blue]{hyperref}

\title{log-can: local-global Class-aware Network for semantic segmentation of remote sensing images}

\name{Xiaowen Ma, Mengting Ma, Chenlu Hu, Zhiyuan Song, Ziyan Zhao, Tian Feng, Wei Zhang \thanks{This work was supported in part by the National Natural Science Foundation of China under Grant 62202421; in part by Zhejiang Provincial Key Research and Development Program under Grant 2021C01031; in part by Ningbo Yongjiang Talent Introduction Programme under Grant 2021A-157-G; and in part by the Public Welfare Science and Technology Plan of Ningbo City under Grant 2022S125.}}

\address{Zhejiang University}

\begin{document}
\maketitle

\begin{abstract}
Remote sensing images are known of having complex backgrounds, high intra-class variance and large variation of scales, which bring challenge to semantic segmentation. We present LoG-CAN, a multi-scale semantic segmentation network with a global class-aware (GCA) module and local class-aware (LCA) modules to remote sensing images. Specifically, the GCA module captures the global representations of class-wise context modeling to circumvent background interference; the LCA modules generate local class representations as intermediate aware elements, indirectly associating pixels with global class representations to reduce variance within a class; and a multi-scale architecture with GCA and LCA modules yields effective segmentation of objects at different scales via cascaded refinement and fusion of features. Through the evaluation on the ISPRS Vaihingen dataset and the ISPRS Potsdam dataset, experimental results indicate that LoG-CAN outperforms the state-of-the-art methods for general semantic segmentation, while significantly reducing network parameters and computation. Code is available at~\href{https://github.com/xwmaxwma/rssegmentation}{https://github.com/xwmaxwma/rssegmentation}.
\end{abstract}


\section{Introduction}
\label{sec:intro}

Semantic segmentation of remote sensing images aims to assign definite classes to each image pixel, which makes important contributions to land use, yield estimation, and resource management~\cite{land1,land2,land4}. Compared to natural images, remote sensing images are coupled with sophisticated characteristics (e.g., complex background, high intra-class variance, and large variation of scales) that potentially challenge the semantic segmentation.

Existing methods of semantic segmentation based on convolutional neural networks (CNN) focus on context modeling ~\cite{pspnet,nonlocal,danet,ccnet}, which can be categorized into spatial context modeling and relational context modeling. Spatial context modeling methods, such as PSPNet ~\cite{pspnet} and DeepLabv3+~\cite{deeplabv3+}, use spatial pyramid pooling (SPP) or atrous spatial pyramid pooling (ASPP) to integrate spatial contextual information. Although these methods can capture the context dependencies with homogeneity, they disregard the differences of classes. Therefore, unreliable contexts may occur when a general semantic segmentation method processes remote sensing images with complex objects and large spectral differences.

Regarding the relational context modeling, non-local neural networks~\cite{nonlocal} compute the pairwise pixel similarities in the image using non-local blocks for weighted aggregation, and DANet~\cite{danet} adopts spatial attention and channel attention for selective aggregation. However, the dense attention operations used by these methods enable a large amount of background noise given the complex background of remote sensing images, leading to the performance degradation in semantic segmentation.
Recent class-wise context modeling methods, such as ACFNet~\cite{acfnet} and OCRNet~\cite{ocrnet}, integrate class-wise contexts by capturing the global class representations to partially prevent the background inference caused by dense attentions. Despite the fact that these methods have achieved ideal performance in semantic segmentation on natural images, the performance on remote sensing images remains problematic, specifically for high intra-class variance that leads to the large gap between pixels and the global class representations. Therefore, introducing local class representations may address this issue.

Given the above observations, we design a global class-aware (GCA) module to capture the global class representations, and local class-aware (LCA) modules to generate the local class representations. In particular, local class representations are used as intermediate aware elements to indirectly associate pixels with global class representations, which alleviates the complex background and the high intra-class variance of remote sensing images. Both modules are integrated into LoG-CAN, a semantic segmentation network with a multi-scale design that improves the large variation of scales issue of remote sensing images.

The primary contributions of this paper are summarized as follows:
\begin{itemize}

\item a novel local class-aware module using the local class representations for class-wise context modeling;

\item a multi-scale semantic segmentation network integrating both local and global class-aware modules;

\item the state-of-the-art performance on two benchmark datasets for aerial images and a significant reduction of the number of parameters and computational efforts.

\end{itemize}

\begin{figure*}[t]
  \centering
  \includegraphics[width=0.77\textwidth]{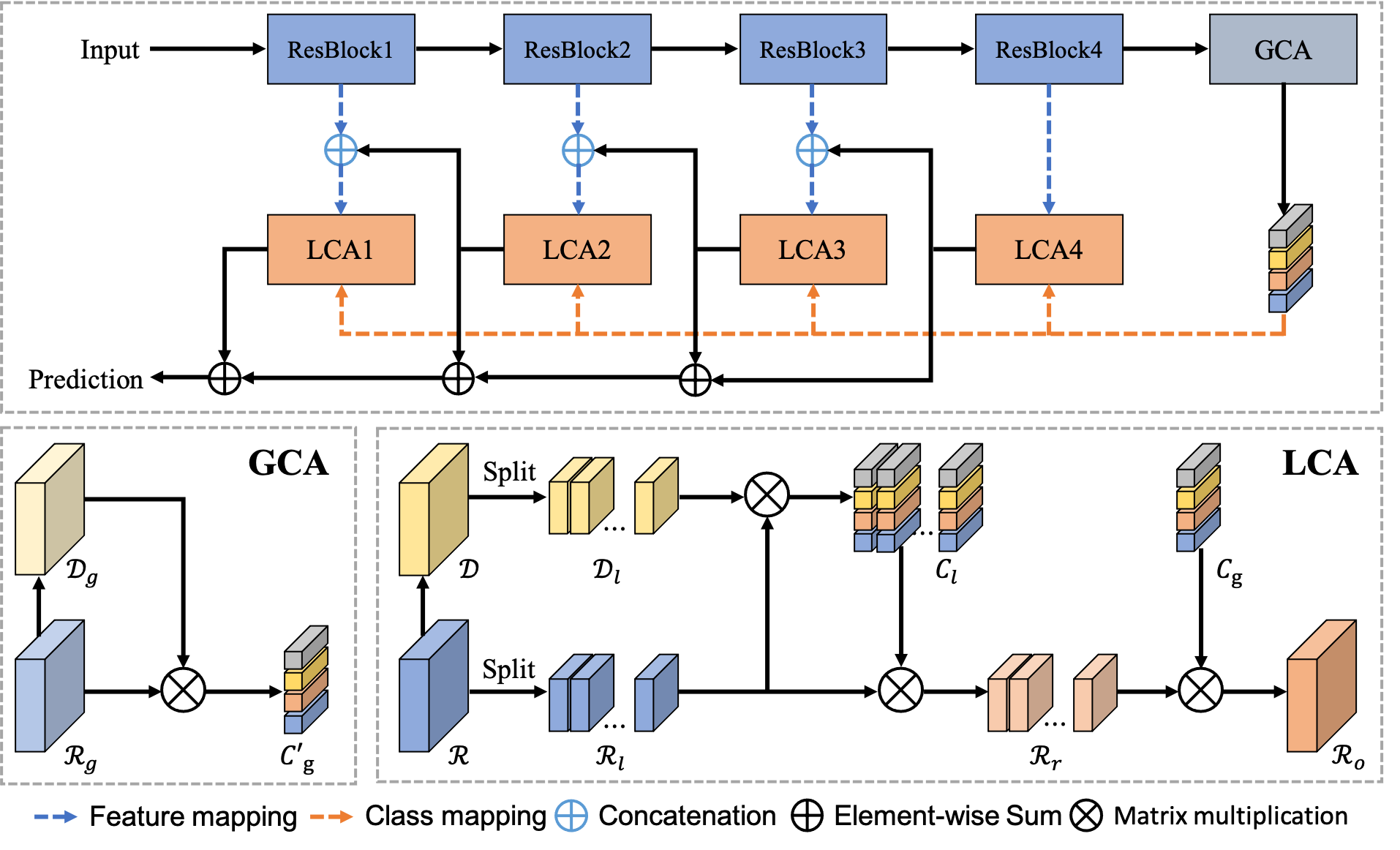}
  \caption{Architecture of LoG-CAN with GCA and LCA modules.}
  \label{fig:whole}
  \vspace{-2mm}
\end{figure*}

\section{Method}
\label{sec:format}

\subsection{Overall Architecture}
The proposed LoG-CAN has an encoder-decoder architecture (as shown in Fig.~\ref{fig:whole}). The encoder uses ResNet50~\cite{resnet} as the backbone for multi-scale feature extraction, and the decoder consists of a global class-aware (GCA) module and local class-aware (LCA) modules to refine multi-scale feature representations from the backbone via class-wise context modeling. Specifically, each residual block $i$ of the four extracts multi-scale feature representations $\mathcal{R}_i$ from the input image; the feature representations $\mathcal{R}_g$ from the last residual block are processed by the GCA module to obtain the intermediate global class representations $\mathcal{C'}_g$. Then, each $\mathcal{R}_i$ and the i+1-th LCA module's output are processed with feature mapping and concatenation to reach intermediate feature representations $\mathcal{R'}$. In addition, the feature representations $\mathcal{R}$ and the class representations $\mathcal{C}_g$ input to the LCA module are obtained via feature mapping and class mapping from $\mathcal{R'}$ and $\mathcal{C'}_g$. Being refined by the cascaded LCA modules, the feature representations at different spatial scales are element-wisely summed and quadruply upsampled for the semantic segmentation output.

Note that our design of feature mapping and class mapping, which are implemented respectively by a $3\times 3$ convolution layer and a $1\times 1$ convolution layer, enables the following two effects: (1) the multi-scale feature representations and class representations further interact with each other in a specific feature space after mapping; (2) mapping reduces the feature channels of both representations, creating a lighter structure that contains fewer model parameters and computation without degrading the model performance.

\subsection{Global Class-Aware Module}
Motivated by~\cite{ocrnet}, we design a GCA module to capture the global class representations. With feature representations $\mathcal{R}_g\in \mathbb{R} ^{C'\times H'\times W'}$ that contain rich semantic information, the distribution of class probability $\mathcal{D}_g$ is obtained as follows,
\begin{equation}
\mathcal{D} _{g} = \mathcal{H} (\mathcal{R}_g ) ,
\end{equation} 
where $\mathcal{D}_g$ is a matrix of size $K\times H'\times W'$ and $K$ is the number of classes. $\mathcal{H}$ is implemented by two consecutive $1\times 1$ convolution layers.
Then, the global class representations $\mathcal{C'}_g$ is defined as follows,
\begin{equation}
\mathcal{C'} _{g} = \mathcal{D} _{g}^{K\times (H'\times W')}\otimes \mathcal{R}_g ^{(H'\times W')\times C'} ,
\end{equation} 
where $\mathcal{C'} _g$ is a matrix of size $K\times C'$.

\setlength{\tabcolsep}{10pt}
\begin{table*}[t]
    \small
	\begin{center}
		\caption{Effectiveness comparison with the state-of-the-art methods on the test set from the ISPRS Vaihingen dataset. Per-class best performance is marked in bold.
				}
		\label{table:1}
		\begin{tabular}{cccccccccc}
		\toprule
			Method &Imp. Sur. &Building &Low Veg. &Tree &Car  &AF &mIoU &OA\\
			\midrule
			PSPNet~\cite{pspnet}  & 91.38 & 94.20 &83.05 &88.71 &75.02 &86.47 &76.78 &89.36\\
			DeepLabv3+~\cite{deeplabv3+} & 91.63 & 94.09 &82.51 &88.00 &77.66 &86.77 &77.13 &89.12\\
			DANet~\cite{danet} & 91.38 & 94.10 &83.09 &89.02 &76.80 &86.88 &77.32 &89.47\\
			Semantic FPN~\cite{semanticfpn} & 91.78 & 94.37 &82.87 &89.44 &79.45 &87.58 &77.94 &89.86 \\
			FarSeg~\cite{farseg} & 92.13 & 94.57 &82.87 &88.74 &81.11 &87.88 &79.14 &89.57 \\
			OCRNet~\cite{ocrnet} & 92.87 & 95.14 &84.32 &89.23 &84.52 &89.22 &81.71 &90.47 \\
			LANet ~\cite{lanet} & 92.41 & 94.90 &82.89 &88.92 &81.31 &88.09 &79.28 &89.83\\
			BoTNet~\cite{botnet} & 92.22 & 94.48 &83.97 &89.57 &82.93 &88.63 &79.89 &90.16\\
			MANet~\cite{manet} & 93.02 & 95.47 &84.64 &89.98 &88.95 &90.41 &82.71 &90.96 \\
			UNetFormer~\cite{unetformer} &92.70 &95.30 &84.90 &90.60 &88.50 &90.40 &82.70 &91.00\\
			\midrule
			LoG-CAN (Ours)          
 &\bf93.71 &\bf96.64 &\bf85.89 &\bf90.93 &\bf90.16 &\bf91.46 &\bf84.13 &\bf91.97 \\
			\bottomrule
		\end{tabular}
	\end{center}
\end{table*}
\setlength{\tabcolsep}{2pt}

\subsection{Local Class-Aware Module}
For remote sensing images, class-wise context modeling that only uses the global class representations circumvents the interference of noise caused by intensive attention operations. However, it can potentially lead to considerable semantic differences between pixels and the global class representations due to the insufficient consideration of high intra-class variance, which degenerates the semantic segmentation performance. In this regard, we exploit the local class representations as an intermediate awareness element to capture the relationship between pixels and the local class representations and aggregate this relationship with the global class representations for class-wise context modeling.

For the feature representations $\mathcal{R}\in \mathbb{R} ^{C\times H\times W}$, we deploy a pre-classification operation for the corresponding distribution $\mathcal{D}\in \mathbb{R} ^{K\times H\times W}$. In particular, we split $\mathcal{R}$ and $\mathcal{D}$ along the spatial dimension to get $\mathcal{R}_l$ and $\mathcal{D}_l$, followed by calculating the local class representations $\mathcal{C}_l$ as follows,
\begin{equation}
\mathcal{C} _{l} = \mathcal{D} _{l}^{(N_h\times N_w) \times K\times ( h \times w)}\otimes \mathcal{R} _{l}^{(N_h\times N_w) \times ( h \times w)\times C} ,
\end{equation} 
where $h$ and $w$ represent the height and width of the selected local patch, $N_h = \frac{H}{h}$, and $N_w = \frac{W}{w}$. The corresponding affinity matrix $\mathcal{R}_r$, which represents the similarity between the pixel and the local class representations, is obtained as follows,
\begin{equation}
\mathcal{R} _{r} = \mathcal{R} _{l}^{(N_h\times N_w) \times ( h \times w)\times C}\otimes \mathcal{C}_l ^{{(N_h\times N_w) \times C\times K}} .
\end{equation} 

Finally, we utilize $\mathcal{R}_r$ to associate the global class representations $\mathcal{C}_g$ and acquire the augmented representations $\mathcal{R}_o$ ,
\begin{equation}
\mathcal{R} _{o} = \psi(\mathcal{R} _{r}^{(N_h\times N_w) \times ( h \times w)\times K}\otimes \mathcal{C}_g ^{K\times C}) ,
\end{equation} 
where $\psi $ is a function that puts the per-local enhanced representations back in place in $\mathcal{R}$.

\section{experiments}
\label{sec:pagestyle}
We implemented the proposed method and evaluated LoG-CAN on the ISPRS Vaihingen dataset and the ISPRS Potsdam dataset using three common metrics: average F1-score (AF), mean Intersection-over-Union (mIoU), and overall accuracy (OA). ISPRS Vaihingen dataset~\cite{www.isprs.org} includes 33 true orthophoto (TOP) tiles and the corresponding digital surface model (DSMs) collected from a small village, where the image size varies from $1996\times 1995$ to $3816\times 2550$ pixels and the ground truth labels comprise six land-cover classes (i.e., impervious surfaces, building, low vegetation, tree, car, and clutter/background). We used 16 images for training and the remaining 17 for testing. ISPRS Potsdam dataset~\cite{www.isprs.org} includes 38 TOP tiles and the corresponding DSMs collected from a historic city with large building blocks.  All images have the same size of $6000\times 6000$ pixels and the ground truth labels comprise the same six land-cover classes as the ISPRS Vaihingen dataset. We used 24 images for training and the remaining 14 for testing.

\subsection{Implementation Details}
We selected ResNet-50~\cite{resnet} pretrained on ImageNet as the backbone for all experiments. The optimizer was SGD with batch size of 8, and the initial learning rate was set to 0.01 with a poly decay strategy and a weight decay of 0.0001.  Following previous work~\cite{lanet,manet}, we randomly cropped the images from both datasets to produce $512\times 512$ patches, and the augmentation methods, such as random scale ([0.5, 0.75, 1.0, 1.25, 1.5]), random vertical flip, random horizontal flip and random rotate, were adopted in the training process. The number of epochs was set to 150 with the ISPRS Vaihingen dataset and 80 with the ISPRS Potsdam dataset.
\setlength{\tabcolsep}{10pt}
\begin{table}[t]
    \small
	\begin{center}
		\caption{
    Effectiveness comparison with the state-of-the-art methods on the test set from the ISPRS Potsdam dataset. Per-class best performance is marked in bold.
		}
		\label{table:2}
		\begin{tabular}{cccc}
		\toprule
			Method &AF &mIoU &OA\\
			\midrule
			PSPNet~\cite{pspnet}  & 89.98 &81.99 &90.14\\
			DeepLabv3+~\cite{deeplabv3+} &90.86 &84.24 &89.18\\
			DANet~\cite{danet}  &89.60 &81.40 &89.73\\
			Semantic FPN~\cite{semanticfpn} & 91.53 &84.57 &90.16\\
			FarSeg~\cite{farseg}  &91.21 & 84.36 &89.87\\
			OCRNet~\cite{ocrnet}  &92.25 &86.14 &90.03\\
			LANet ~\cite{lanet}  &91.95 &85.15 &90.84\\
			BoTNet~\cite{botnet}  &91.77 &84.97 &90.42\\
			MANet~\cite{manet}  &92.90 &86.95 &91.32\\
			UNetFormer~\cite{unetformer}  &92.80 &86.80 &91.30\\
			\midrule
			LoG-CAN (Ours) &\bf93.53 &\bf87.69 &\bf92.09 \\
			\bottomrule
		\end{tabular}
	\end{center}
\end{table}
\setlength{\tabcolsep}{2pt}

\begin{figure}[t]
\vspace{-5mm}
\centering
\captionsetup[subfloat]{labelsep=none,format=plain,labelformat=empty}
\subfloat[Input]{
\begin{minipage}[t]{0.18\linewidth}
\includegraphics[width=1\linewidth]{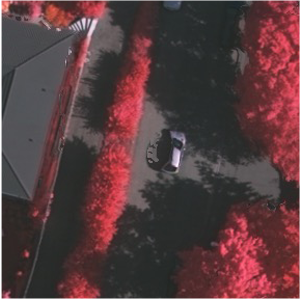}\vspace{4pt}
\includegraphics[width=1\linewidth]{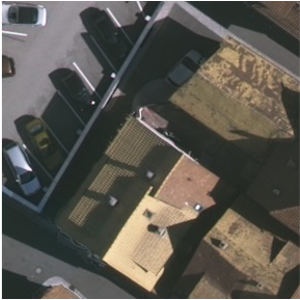}\vspace{4pt}
\end{minipage}}
\subfloat[GT]{
\begin{minipage}[t]{0.18\linewidth}
\includegraphics[width=1\linewidth]{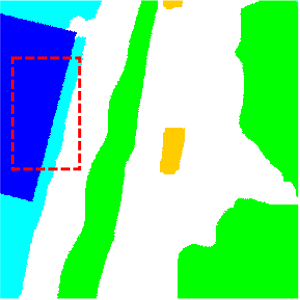}\vspace{4pt}
\includegraphics[width=1\linewidth]{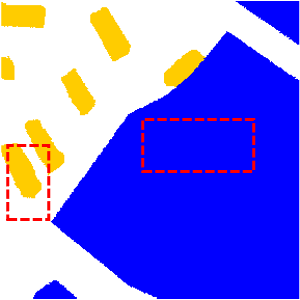}\vspace{4pt}
\end{minipage}}
\subfloat[PSPNet]{
\begin{minipage}[t]{0.18\linewidth}
\includegraphics[width=1\linewidth]{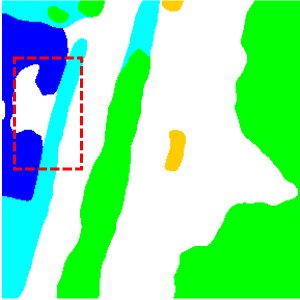}\vspace{4pt}
\includegraphics[width=1\linewidth]{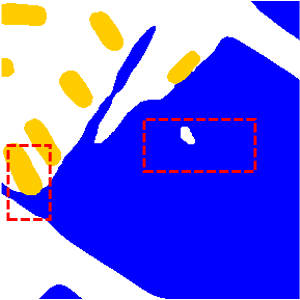}\vspace{4pt}
\end{minipage}}
\subfloat[MANet]{
\begin{minipage}[t]{0.18\linewidth}
\includegraphics[width=1\linewidth]{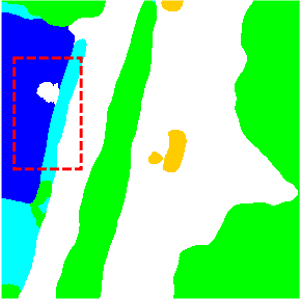}\vspace{4pt}
\includegraphics[width=1\linewidth]{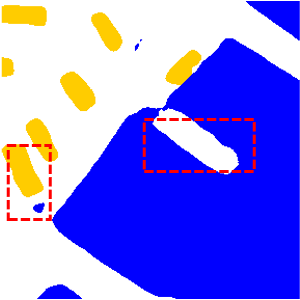}\vspace{4pt}
\end{minipage}}
\subfloat[LoG-CAN]{
\begin{minipage}[t]{0.18\linewidth}
\includegraphics[width=1\linewidth]{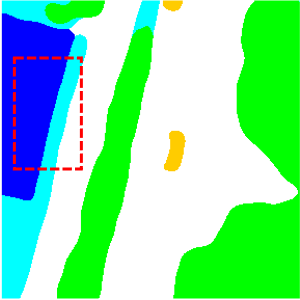}\vspace{4pt}
\includegraphics[width=1\linewidth]{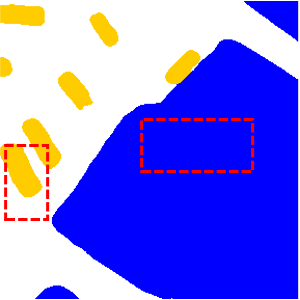}\vspace{4pt}
\end{minipage}}
\caption{Example outputs from the LoG-CAN and other methods on the ISPRS Vaihingen dataset. Best viewed in color and zoom in.}
\label{fig:color}
\end{figure}

\setlength{\tabcolsep}{4pt}
\begin{table}[t]
    \small
	\begin{center}
		\caption{Computational complexity comparison with other popular context aggregation modules. Per-class best performance is marked in bold.
		}
		\label{table:3}
		\begin{tabular}{cccc}
		\toprule
			Method &Params (M) &FLOPs (G) &Memory (MB)\\
			\midrule
			PPM~\cite{pspnet} & 23.1 & 309.5 & 257 \\
			ASPP~\cite{deeplabv3+} & 15.1 & 503.0 & 284 \\
			DAB~\cite{danet} &23.9 & 392.2 & 1546\\
			OCR~\cite{ocrnet} &10.5 & 354.0 &202 \\
			PAM+AEM~\cite{lanet} &10.4 & 157.6 &489\\
			ILCM+SLCM~\cite{isnet} &11.0 &180.6 &638\\
			KAM~\cite{manet} &5.3 &85.9 &160\\
			
			\midrule
			LCA (Ours) &\bf0.8 &\bf11.9 &\bf53\\
			\bottomrule
		\end{tabular}
	\end{center}
\end{table}
\setlength{\tabcolsep}{2pt}

\begin{figure}[t]
  \vspace{-5mm}
  \centering
  \includegraphics[width=0.5\textwidth]{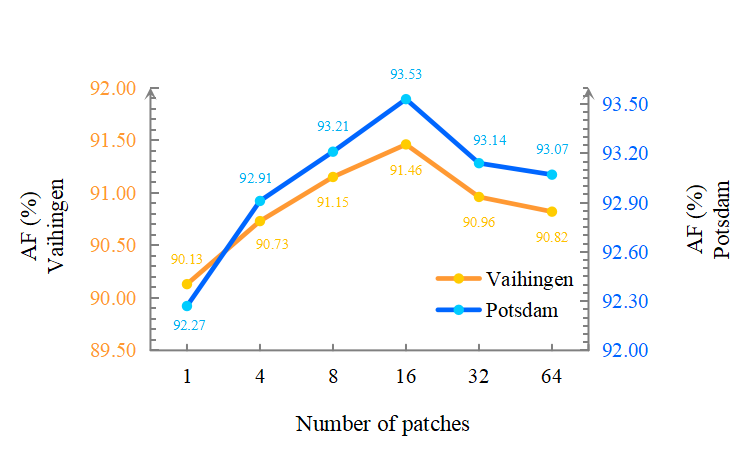}
  \caption{ Plot of AF against the number of patches on the ISPRS Vaihingen dataset (yellow) and the ISPRS Potsdam dataset (blue) }
  \label{fig:2}
\end{figure}

\subsection{Evaluation and Analysis}
As shown in Table~\ref{table:1}, the proposed method outperformed other state-of-the-art methods on the ISPRS Vaihingen dataset in AF, mIoU, and OA. In particular, our LoG-CAN achieved the AF of 91.46\% and the mIoU of 84.13\%, even higher than MANet~\cite{manet} and UNetFormer~\cite{unetformer}, showing that our design on class-wise context modeling has greater effectiveness. As shown in Table~\ref{table:2}, our LoG-CAN also reached outstanding performances in all metrics on the ISPRS Potsdam dataset. Fig.~\ref{fig:color} shows example result outputs from our 
LoG-CAN, PSPNet, and MANet. In particular, the proposed method not only better preserves the integrity and regularity of semantic objects, but also improves the segmentation performance of small objects.

To validate the lightness of our method, we compare our LCA module with several classical context aggregation modules, including the number of parameters measured in million (M), the floating-point operations per second (FLOPs) measured in giga (G), and the memory consumption measured in megabytes (MB). All inputs were set to the size of $2048\times 128\times 128$ to ensure the comparison's fairness. As shown in Table~\ref{table:3}, the LCA module enables significantly less number of parameters and lower computation compared to PPM~\cite{pspnet}. From the perspective of the 
entire network's structure, our LoG-CAN only needs 60\% of the parameters and 25\% of the GFLOPs compared to PSPNet~\cite{pspnet}, which suggests its design as a lightweight method.

We investigated if the number of patches in the LCA module has any impact on the results. As shown in Figure~\ref{fig:2}, the best result was obtained on each dataset with the number of patches being set to 16. Besides, when the number of patches was set to 1, the local class representations degenerated to the global class representations, resulting into relatively unsatisfactory performances. These findings indicate that local class awareness can effectively improve class-wise context modeling.

\section{conclusion}
\label{sec:typestyle}
In this paper, we introduce LoG-CAN for semantic segmentation of remote sensing images. Our method effectively resolves the problems due to complex background, high intra-class variance, and large variation of scales in remote sensing images by combining the global and local class representations for class-wise context modeling with a multi-scale design. According to the experimental results, LoG-CAN has greater effectiveness than the state-of-the-art general methods for semantic segmentation, while requiring less network parameters and computation. The proposed method provides a better trade-off between efficiency and accuracy.

\bibliographystyle{IEEEbib}
\bibliography{refs}

\end{document}